%% file: main.tex
\documentclass[11pt]{article}

\usepackage{acl}

\usepackage{times}
\usepackage{latexsym}

\usepackage[T1]{fontenc}

\usepackage[utf8]{inputenc}

\usepackage{microtype}

\usepackage{inconsolata}

\usepackage{graphicx}
\usepackage{hyperref}
\usepackage{url}

\usepackage{caption}
\usepackage{enumitem}
\usepackage{multirow}
\usepackage{makecell}
\usepackage{subfigure}
\usepackage{booktabs}
\usepackage{amsmath}
\usepackage{amssymb}
\usepackage{tabularx}
\usepackage{makecell}

\title{HeTGB: A Comprehensive Benchmark for Heterophilic Text-Attributed Graphs}

\author{
    Shujie Li\textsuperscript{1}$^{*}$,
    Yuxia Wu\textsuperscript{2}$^{*}$,
    Yuan Fang\textsuperscript{2},
    Chuan Shi\textsuperscript{1}
    \\
    \textsuperscript{1}Beijing University of Posts and Telecommunications,
    \textsuperscript{2}Singapore Management University\\
    \text{shujieli@bupt.edu.cn, yieshah2017@gmail.com},
    \text{yfang@smu.edu.sg, shichuan@bupt.edu.cn}
}

\begin{document}
\maketitle
\begingroup
\renewcommand\thefootnote{\relax}
\footnotetext{$^{*}$ Co-first authors with equal contribution.}
\endgroup
\begin{abstract}
Graph neural networks (GNNs) have demonstrated success in modeling relational data primarily under the assumption of homophily. However, many real-world graphs exhibit heterophily, where linked nodes belong to different categories or possess diverse attributes, such as webpages, Wikipedia articles, social networks, and e-commerce platforms. Additionally, nodes in many domains are associated with textual descriptions, forming heterophilic text-attributed graphs (TAGs). Despite their significance, heterophilic TAGs remain underexplored due to the lack of dedicated benchmarks that jointly capture heterophilic structures and rich textual attributes. To address this gap, we introduce the \textbf{He}terophilic \textbf{T}ext-attributed \textbf{G}raph \textbf{B}enchmark (HeTGB), a novel benchmark comprising five real-world heterophilic graph datasets from diverse domains, with nodes enriched by extensive textual descriptions. HeTGB enables systematic evaluation of GNNs, pre-trained language models (PLMs) and co-training methods on the node classification task. Through extensive benchmarking experiments, we showcase the utility of text attributes in heterophilic graphs, analyze the challenges posed by heterophilic TAGs and the limitations of existing models, and provide insights into the interplay between graph structures and textual attributes.\footnote{All datasets and baselines are available at \url{https://github.com/honey0219/HeTGB}.}
\end{abstract}

\input{intro}
\input{related}
\input{method}

\input{exp}

\section{Conclusion}
In this work, we introduced HeTGB, a dedicated benchmark for heterophilic text-attributed graphs, addressing the limitations of existing datasets and evaluation frameworks. We collected and provided five datasets from different domains, ensuring high-quality textual content and diverse heterophilic structures. Using these datasets, we systematically investigate the performance of prevailing learning paradigms, including GNN-based, PLM-based, and co-training methods, analyzing their strengths and limitations across different heterophilic datasets. Our experimental results validate the effectiveness of HeTGB as a benchmark and offer valuable insights to guide future research and advancements in heterophilic graph learning.

\section{Limitations}

While HeTGB provides diverse heterophilic TAG datasets and a unified evaluation framework, several limitations remain. First, the current benchmark focuses primarily on node classification and does not yet cover other relevant tasks such as link prediction, graph classification, and anomaly detection. Second, only a limited number of large-scale heterophilic TAGs are currently included, which may constrain evaluation at larger scales. Extending HeTGB to broader task settings and incorporating additional large-scale datasets are important directions for future work.

\bibliography{ref}

\appendix
\input{app}

\end{document}

%% file: intro.tex
\section{Introduction}

Graphs serve as a foundational data structure for representing relational information across diverse domains, including webpages, Wikipedia articles, social networks, citation networks, and knowledge graphs \cite{wu2020comprehensive}. 
Conventional graph neural networks (GNNs) predominantly rely on the assumption of \emph{homophily}, meaning that neighboring nodes on the graph share similar characteristics \cite{kipf2016semi,velivckovic2018graph}. However, numerous real-world graphs exhibit \textit{heterophily}, in which linked nodes belong to distinct classes or possess dissimilar attributes \cite{zheng2022graph}.  In such heterophilic graphs, nodes often contain textual attributes that provide rich semantic information, forming heterophilic text-attributed graphs (TAGs). For example, in the Wikipedia hyperlink graph, articles on diverse topics may link to one another, yet their textual content provides essential context for understanding these connections. However, the data and methodologies tailored for heterophilic TAGs remain underexplored, necessitating a systematic investigation into their unique challenges, requirements and insights. Consequently, there is an increasing demand for benchmarks that jointly evaluate structural and textual modeling under heterophily.

While prior works on heterophilic graph learning have introduced various datasets spanning different domains \cite{peigeom, lim2021large, platonovcritical}, these datasets only contain categorical features or shallow features derived from raw attributes, lacking the rich textual descriptions that could provide deeper semantic insights. Meanwhile, existing approaches to heterophilic graphs mainly focus on GNN-based architectures designed to capture structural patterns, incorporating non-local neighbor expansion or architectural refinements \cite{zheng2022graph, gong2024towards}. Notably, these methods do not explicitly integrate textual attributes, limiting their effectiveness in domains where semantic content is crucial.
On the other hand, the remarkable advancements of pre-trained language models (PLMs)
in text comprehension and generation---ranging from smaller models like BERT \cite{kenton2019bert} to large language models (LLMs)---have driven growing interest in methodologies for TAGs. For instance, the recently proposed CS-TAG benchmark \cite{yan2023comprehensive} provides a comprehensive analysis by incorporating various learning paradigms.
While recent studies explore hybrid architectures that integrate PLMs and GNNs in a unified pipeline \cite{liuone, tang2024graphgpt}, their evaluations remain largely confined to homophilic graphs.

Despite recent efforts for both heterophilic graphs and TAGs, a significant research gap persists in two key aspects. (1) \textit{Lack of dedicated heterophilic TAG datasets}: Current heterophilic graph datasets predominantly focus on structural properties while lacking textual content, limiting their ability to evaluate PLM-based and co-training methods. Conversely, most TAG datasets exhibit homophilic connectivity, restricting their relevance for assessing models designed for heterophilic graphs. (2) \textit{Absence of systematic analysis of existing methodologies for heterophilic TAGs.} While many methods exist for heterophilic graphs, it remains unclear whether they can effectively leverage rich textual information. Furthermore, it is uncertain whether state-of-the-art methods for TAGs are sufficiently robust for heterophilic graphs.

To address the above two aspects, we introduce \textbf{He}terophilic \textbf{T}ext-attributed \textbf{G}raph \textbf{B}enchmark (HeTGB), which comprises five real-world heterophilic graph datasets from diverse domains, with nodes enriched by extensive textual descriptions. To construct these datasets, we carefully select datasets from various domains and extract their raw texts from publicly accessible data sources. With HeTGB, we provide a systematic evaluation of representative GNN-based, PLM-based and co-training methods on the node classification task. The experimental analysis highlights the challenges and limitations of current algorithms, offering valuable insights into future research directions. 

Our main contributions are summarized as follows.
(1) We introduce HeTGB, a novel benchmark that integrates heterophilic graph structures with rich textual attributes, addressing the gap left by both existing heterophilic graph benchmarks and text-attributed graph benchmarks.
(2) We conduct a systematic evaluation of representative GNN-based, PLM-based and co-training models on the node classification task, identifying key challenges, limitations and insights in heterophilic TAG learning.
(3) We publicly release HeTGB with datasets and baseline implementations, to facilitate further research and encourage the development of effective models for heterophilic TAGs.

%% file: related.tex
\section{Related Work}

\vspace{+0.1cm}\noindent\textbf{Heterophilic graph learning.}
The widely used benchmarks for heterophilic graphs originate from \cite{peigeom}, including webpages such as \textit{Texas}, \textit{Cornell}, and \textit{Wisconsin}. Subsequently, \citet{lim2021large} and \citet{platonovcritical} introduced larger and more structurally diverse datasets. More recently, \citet{lin2025heterophily} proposed a large-scale benchmark that considers both heterophily and heterogeneity, further enriching the diversity of graph structures studied in this area. However, these benchmarks predominantly lack rich textual attributes, with node features typically comprising categorical or numerical values, or shallow text representations like bag-of-words or FastText embeddings \cite{grave2018learning}, limiting their relevance for text-rich scenarios. On the methods side, GNN-based approaches to heterophily fall into two main categories: neighborhood expansion and architectural refinement. The former enhances expressiveness by incorporating high-order neighborhood aggregation \cite{songordered, yu2024non} or global graph-based identification \cite{wang2022powerful2, bi2024make}, while the latter improves message propagation by adaptively aggregating information from homophilic and heterophilic neighbors \cite{bo2021beyond, liang2024predicting} or leveraging inter-layer interactions to capture multi-hop dependency \cite{chienadaptive}.

\vspace{+0.1cm}\noindent\textbf{Text-attributed graph learning.}
TAGs integrate textual content with graph structure, making them suitable for tasks in document classification, citation networks, and knowledge graphs. While benchmarks such as CS-TAG \cite{yan2023comprehensive}, dynamic TAGs \cite{zhang2024dtgb}, and textual-edge graphs \cite{li2024teg} have advanced evaluation in this space, a dedicated benchmark specifically for heterophilic TAGs remains unexplored.
On the modeling side, early efforts enhanced node features with static word embeddings, evolving to contextual embeddings using SLMs like BERT \cite{kenton2019bert} and later to LLMs such as GPT-4 and LLaMA \cite{achiam2023gpt, touvron2023llama}. Integration of PLMs into graph learning has followed two main approaches: PLM-based methods that transform graph data into natural language \cite{zhao2023gimlet, ye2024language, liu2023evaluating}, and co-training approaches that harness the strengths of both paradigms  \cite{heharnessing, tang2024graphgpt, zhang2024graphtranslator}. Despite these advances, most existing works primarily focus on homophilic graphs. HeTGB fills this gap by offering a benchmark designed to evaluate the existing methods for heterophilic TAGs. 
Notably, LLM4HeG \cite{wu2024exploring} represents an early attempt to apply PLMs to heterophilic graphs, highlighting the limited exploration in this direction and further emphasizing the need for HeTGB.

%% file: method.tex
\begin{table*}[t]
\centering
\small
\setlength{\tabcolsep}{2pt}
\caption{Dataset statistics and comparison with existing datasets.}
\label{tab: dataset}
\resizebox{\textwidth}{!}{
\begin{tabular}{lcccccccc}
\hline
Dataset & \#Class & \#Node & \#Edge & Domain & Feature & $\mathcal{H}_{e}(G)$ & Hetero. & Text \\
\hline
\multicolumn{9}{c}{\textit{ (a) Previous Heterophilic Graphs}} \\
\hline
Cornell \cite{peigeom}          & 5 & 183     & 295      & Web page   & BOW & 0.30 & \checkmark & $\times$ \\
Texas \cite{peigeom}            & 5 & 183     & 309      & Web page   & BOW & 0.06 & \checkmark & $\times$ \\
Wisconsin \cite{peigeom}        & 5 & 251     & 499      & Web page   & BOW & 0.17 & \checkmark & $\times$ \\
Actor \cite{peigeom}            & 5 & 7600    & 33544    & Social     & Keywords     & 0.22 & \checkmark & $\times$ \\
Amazon \cite{platonovcritical}           & 5 & 24492   & 93050    & E-commerce & FastText     & 0.38 & \checkmark & $\times$ \\
Roman-empire \cite{platonovcritical}     & 18 & 22662  & 32927    & Wikipedia  & FastText     & 0.05 & \checkmark & $\times$ \\
Tolokers \cite{platonovcritical}         & 2 & 11758   & 519000   & Crowdsourcing      & Numerical    & 0.59 & \checkmark & $\times$ \\
\hline
\multicolumn{9}{c}{\textit{(b) Previous Text-attributed Graphs}} \\
\hline
Cora \cite{liuone}             & 7 & 2708    & 5429     & Academic   & PLMs         & 0.81 & $\times$   & \checkmark \\
Pubmed \cite{liuone}           & 3 & 19717   & 44338    & Academic   & PLMs         & 0.80 & $\times$   & \checkmark \\
Arxiv \cite{yan2023comprehensive}            & 40 & 169343 & 1166243  & Academic   & PLMs         & 0.66 & $\times$   & \checkmark \\
History \cite{yan2023comprehensive}          & 12 & 41551  & 358574   & E-commerce & PLMs         & 0.66 & $\times$   & \checkmark \\
Computers \cite{yan2023comprehensive}        & 10 & 97229  & 721080   & E-commerce & PLMs         & 0.83 & $\times$   & \checkmark \\
Sports \cite{yan2023comprehensive}           & 13 & 173055 & 1773500  & E-commerce & PLMs         & 0.90 & $\times$   & \checkmark \\
Children \cite{yan2023comprehensive}$^{*}$         & 24 & 76875  & 1554578  & E-commerce & PLMs         & 0.42 & \checkmark & \checkmark \\
\hline
\multicolumn{9}{c}{\textit{ \textbf{(c) Our HeTGB}}} \\
\hline
Cornell   & 5 & 195     & 304      & Web page   & PLMs         & 0.13 & \checkmark & \checkmark \\
Texas     & 5 & 187     & 328      & Web page   & PLMs         & 0.12 & \checkmark & \checkmark \\
Wisconsin & 5 & 265     & 530      & Web page   & PLMs         & 0.20 & \checkmark & \checkmark \\
Actor     & 5 & 4416    & 12172    & Social     & PLMs         & 0.56 & \checkmark & \checkmark \\
Amazon    & 5 & 24492   & 93050    & E-commerce & PLMs         & 0.38 & \checkmark & \checkmark \\
\hline

    
\end{tabular}
}

\captionsetup{justification=raggedright, singlelinecheck=false}
\caption*{\scriptsize $^{*}$ The Children \cite{yan2023comprehensive} exhibits a notably low homophily ratio. Accordingly, we provide additional experiments in Appendix \ref{app:largescale}.}
\vspace{-0.6cm}
\end{table*}

\section{HeTGB: A Comprehensive Benchmark for Heterophilic TAGs}
In this section, we begin with an overview of HeTGB, followed by detailed descriptions of the dataset construction.

\subsection{Overview of HeTGB}
HeTGB is introduced to address the absence of standardized benchmarks for heterophilic TAGs. As outlined in Table~\ref{tab: dataset}(c), HeTGB includes five real-world datasets from diverse domains, ranging from Web pages to e-commerce and social networks. A comparison of HeTGB to previous heterophilic graph benchmarks and text-attributed graphs is shown in  Table~\ref{tab: dataset}(a) and (b), respectively. In these tables, $\mathcal{H}_{\text{e}}$ denotes the edge homophily ratio. Detailed definitions, dataset statistics, and distribution analyses are provided in Appendix~\ref{app:distribution}.



HeTGB facilitates the systematic evaluation of graph learning models including GNNs, PLMs, and co-training methods on the supervised node classification tasks. Node classification represents the most fundamental task in heterophilic graph research \cite{luan2022revisiting, zheng2022graph}, as heterophily is primarily reflected in the interaction between the structural connectivity and the distribution of node class labels.
Besides, the benchmark provides standardized data splits, baseline implementations, and performance metrics to ensure reproducible research and foster innovation in heterophilic TAG learning.

\subsection{Dataset Construction}
To construct HeTGB, we carefully selected five real-world datasets, which are characterized by classic heterophilic structures and associated with rich textual content. The construction process involved sourcing publicly available graph structures 
and extracting raw textual content from associated metadata.
Each dataset was preprocessed into a standardized format with pre-defined splits for training, validation, and testing to facilitate fair comparisons. 
Note that, since most of the publicly preprocessed datasets used in existing heterophilic graph work lack raw textual content and details of their preprocessing steps, we reconstructed the datasets from their original data sources. As a result, the number of nodes and edges may slightly differ from their preprocessed counterparts. In what follows, we describe the construction details for each dataset.

\vspace{+0.1cm}\noindent\textbf{Cornell, Texas, and Wisconsin.} They are widely used subsets of the WebKB dataset\footnote{https://www.cs.cmu.edu/afs/cs.cmu.edu/project/theo-11/www/wwkb/} collected by Carnegie Mellon University from computer science departments at various universities. Each node corresponds to a Web page, and edges represent hyperlinks between pages. Each node is categorized into seven categories: student, faculty, staff, department, course, project, and other. These graphs were first utilized for heterophilic graph learning by \citet{peigeom}. 
We obtained a subset of five webpage categories and their hyperlinks from a public repository\footnote{https://github.com/sqrhussain/structure-in-gnn/tree/master/data/graphs/raw/WebKB} \cite{hussain2021impact}. Since only URLs were provided, we matched them to the original WebKB HTML files\footnotemark[1] and extracted the full text to reconstruct the dataset. Compared to the preprocessed counterparts \cite{peigeom} summarized in Table~\ref{tab: dataset}(a), the categories of Web pages are the same, namely, student, project, course, staff, and faculty, while the numbers of nodes and edges remain similar as shown in Table~\ref{tab: dataset}(c). 
However, we observe differences in the homophily ratios compared to previous preprocessed versions \cite{peigeom}. This discrepancy arises due to differences in the node and edge sets, as the dataset provided by \citet{peigeom} only contains preprocessed node IDs without the full textual content. Consequently, our reconstructed graphs do not correspond one-to-one with theirs, resulting in variations in graph topology and homophily ratios.

\vspace{+0.1cm}\noindent\textbf{Actor.} This is an actor-only induced subgraph derived from a film-director-actor-writer network\footnote{https://www.aminer.cn/lab-datasets/soinf/}, curated by  \citet{tang2009social}. The dataset was first used for heterophilic graph learning \cite{peigeom} without raw text, category names and other preprocessing details. Thus, we rely on the original data source \cite{tang2009social} crawled from Wikipedia to extract the actor-only subgraph, in which nodes represent actors, and an edge between two nodes indicates their co-occurrence on the same Wikipedia page. The raw text of each node contains the node ID, the number of words introducing the node on Wikipedia, and the multiple categories extracted from Wikipedia pages. The original source \cite{tang2009social} shows notable category imbalance, with around 66\% of actors labeled as ``\textit{American film actors}'', and many actors belonging to multiple categories such as ``\textit{American film actors}'' and ``\textit{American television actors}''. Inspired by prior preprocessing steps \cite{tang2009social}, we selected actors with high occurrence frequencies and combined related categories to improve balance. Finally, following prior work \cite{peigeom}, we selected five categories, as follows.
\begin{itemize}[noitemsep,topsep=0pt,left=0pt]
    \item ``American film actors (only)'': actors exclusively labeled as ``\textit{American film actors}'';
    \item ``American film actors and American television actors'': category including both ``\textit{American film actors}'' and ``\textit{American television actors}'';
    \item ``American television actors and American stage actors'': category including both ``\textit{American stage actors}'' and ``\textit{American television actors}'';
    \item ``English actors'': category including ``English'' such as ``\textit{English film actors}'', ``\textit{English television actors}'' or ``\textit{English stage actors}'';
    \item ``Canadian actors'': category including  ``\textit{Canadian}''.
\end{itemize}
We removed category terms corresponding to the classification labels from the raw text to prevent label leakage in our classification task. Afterward, we constructed the graph based on the nodes in the five categories and the edges between them and removed isolated nodes from the graph. 

Note that our graph has a higher homophily ratio than the previous version \cite{peigeom} due to differences in category and graph construction. Ultimately, this is still a form of social network, where actors in the same category co-occur on the same Wikipedia page more frequently, increasing homophily. However, it is still considered heterophilic compared to classic homophilic graphs.

\vspace{+0.1cm}\noindent\textbf{Amazon.} It is constructed from the Amazon product co-purchasing network metadata\footnote{https://snap.stanford.edu/data/amazon-meta.html} from Stanford Large Network Dataset Collection \cite{jure2014snap}. Nodes represent products such as books, music CDs, DVDs, and VHS tapes, while edges link products frequently bought together. The classification task is to predict the average product rating, grouped into five classes. This dataset was first used for heterophilic graph learning by \citet{platonovcritical}, where preprocessed data included raw textual descriptions such as product name, category, and rating information. However, their rating information includes only metadata (\textit{e.g.}, average rating, time, user ID and corresponding rating). 
We removed the average rating to prevent label leakage in the classification task. Instead, we retained up to five earliest individual user ratings per product, along with product descriptions, to synthesize the textual content. Following prior work \cite{platonovcritical}, only the largest connected component of the five-core subgraph is considered.


\section{Learning Paradigms for HeTGB} 
In this section, we introduce the existing learning paradigms for heterophilic graphs and TAGs including GNN-based methods, PLM-based methods, and co-training methods.

\subsection{GNN-based Methods} 

GNN-based methods focus on learning from graph structures. However, they have limited capability in handling textual content, often relying on bag-of-words features or shallow embeddings. Both classic and heterophily-specific GNNs have been proposed, as we elaborate in the following.

\vspace{+0.1cm}\noindent\textbf{Classic GNNs.} GNNs typically adopt the message-passing mechanism, where each node representation is updated by aggregating the messages from local neighbors and then combining the aggregated messages with its ego representation \cite{xu2018powerful}. This updating process can be stacked over multiple layers to capture higher-order dependencies. Specifically, 
the representation of node $v \in \mathcal{V}$ in the $l$-th layer can be obtained by: 
\begin{align}
\mathbf{r}_v^{(l)} &= \textsc{Aggr}^{(l)}\left(\{\mathbf{h}_u^{(l-1)} : u \in \mathcal{N}(v)\}\right), \\ \label{eq:gnn:aggr}
\mathbf{h}_v^{(l)} &= \textsc{Combine}^{(l)}\left(\mathbf{h}_v^{(l-1)}, \mathbf{r}_v^{(l)}\right), 
\end{align} 
where $\mathcal{N}(v)$ represents the neighbors of node $v$, and $\mathbf{r}_v^{(l)}$ and $\mathbf{h}_v^{(l)}$ stand for the message vector and the representation vector of node $v$ at the $l$-th layer, respectively. \textsc{Aggr}$(\cdot)$ is an aggregation function (\textit{e.g.}, mean and max pooling), and \textsc{Combine}$(\cdot)$ is a combination function (\textit{e.g.}, a linear layer) \cite{hamilton2017inductive}. 

Note that the general message-passing mechanism operates under the assumption of homophily, where neighbors are assumed to have similar attributes or representations as the ego node.

\vspace{+0.1cm}\noindent\textbf{Heterophily-specific GNNs.} Unlike classic GNNs, heterophily-specific GNNs are designed to adapt their message passing process to heterophilic graphs.  They mainly focus on customizing the neighborhood aggregation and feature updating schemes that specifically model the heterophily characteristics  \cite{zheng2022graph}. Heterophily-specific GNNs can be broadly categorized into non-local neighbor extension methods \cite{songordered,peigeom}, architecture refinement methods \cite{bo2021beyond, du2022gbk,xu2018representation}, and hybrid methods \cite{zhu2020beyond}. 

Non-local neighbor extension methods aim to redefine the neighborhood $\mathcal{N}(v)$ in Eq.~\eqref{eq:gnn:aggr}. Various techniques have been proposed, such as adopting high-order neighbor mixing \cite{abu2019mixhop, songordered}, and potential neighbor discovery based on various distance measures, $\mathcal{N}_p(v) = \{u:\textsc{Dist}(u,v) \leq \delta\}$ \cite{peigeom}. 

Architecture refinement methods focus on revisiting the \textsc{Aggr}$(\cdot)$ and \textsc{Combine}$(\cdot)$ functions tailored to heterophilic graphs. One line of work adopts adaptive message aggregation by assigning different weights for homophilic and heterophilic neighbors \cite{bo2021beyond, du2022gbk}:
\begin{align}
\mathbf{r}_v^{(l)} = \textsc{Aggr}^{(l)}\left(\{\mathbf{\alpha}_{uv}^{(l)}\mathbf{h}_u^{(l-1)} : u \in \mathcal{N}(v)\}\right),
\end{align}
where $\mathbf{\alpha}_{uv}^{(l)}$ denotes the aggregation weight for node pair $(u,v)$ at the $l$-th layer.
Another line of work explores the inter-layer combination to integrate the local information of shallow layers and the global information from deeper layers \cite{xu2018representation}: 
\begin{equation}
\mathbf{h}_v=\textsc{Combine}(\mathbf{h}_v^{(1)},\mathbf{h}_v^{(2)},\cdots,\mathbf{h}_v^{(L)}).
\end{equation}

\subsection{PLM-based Methods}
\label{method-plm}

PLM-based methods harness the advanced language modeling capabilities of PLMs to effectively capture the semantics of textual content on nodes, enabling direct predictions across a broad spectrum of graph-related tasks. Nevertheless, they generally struggle to capture graph structures effectively. A critical component of these methods is transforming graph-structured data into textual descriptions that PLMs can process \cite{li2023survey,liu2023towards}. This is typically achieved through carefully designed prompts that encapsulate the task description, node and edge lists, and textual content of each node. However, in heterophilic graphs, prompts that incorporate structural signals (\textit{e.g.,} summaries of neighboring nodes) may introduce conflicting or misleading information due to the heterophilous nature of neighbors \cite{chen2024exploring}. This underscores the need for more principled prompt designs that selectively integrate structural information when adapting PLM-based methods to heterophilic settings. 

Generally, given a graph $\mathcal{G}=(\mathcal{V},\mathcal{E},\mathcal{T})$ with a set of nodes $\mathcal{V}$, edges $\mathcal{E}$ and textual attributes $\mathcal{T}$, the PLM-based paradigm is formulated as follows: 
\begin{align}
G_{\text{seq}} &= \textsc{Flatten}(\mathcal{V},\mathcal{E},\mathcal{T}), \\
Y &= f_\text{PLM}(\textsc{Prompt}, G_{\text{seq}}),
\end{align}
where $\textsc{Flatten}(\cdot)$ is the flatten function to transform the graph $\mathcal{G}$ into a language sequence. The prediction is then obtained by feeding the prompt and flattened graph sequence $G_{\text{seq}}$ into a PLM.

\subsection{Co-training Methods}

Co-training methods aim to integrate the advantages of GNNs for structural representation learning with the powerful semantic modeling capabilities of PLMs. These methods can be further categorized into GNN-centric, symmetric, and PLM-centric approaches \cite{liu2023towards}, depending on how they couple the GNN and PLMs. 

\vspace{+0.1cm}\noindent\textbf{GNN-centric methods.} They leverage PLMs to extract more representative textual features for nodes, while GNNs are primarily responsible for learning graph representations and making predictions \cite{liuone,wu2024exploring}, as follows. 
\begin{equation}
    Y = f_\text{GNN}(f_\text{PLM}(\mathcal{T}), \mathcal{G}),
\end{equation} 
 where $f_\text{PLM}(\mathcal{T})$ extracts semantic representations from the textual attributes of each node, which are subsequently integrated into a GNN for downstream prediction.
 
\vspace{+0.1cm}\noindent\textbf{Symmetric methods.} They align the structural embeddings generated by GNNs with the semantic ones obtained from PLMs, ensuring consistency between the two modalities \cite{wen2023augmenting}: 
\begin{equation}
 \min \mathcal{L}(f_\text{GNN}(\mathcal{G}), f_\text{PLM}(\mathcal{T})),
 \end{equation} 
 where the objective is to minimize the discrepancy between structure- and text-based embeddings.
 
\vspace{+0.1cm}\noindent\textbf{PLM-centric methods.}  They utilize GNNs to enhance the performance of PLMs. Recent advancements in this area focus on injecting knowledge of complex graph structures into PLMs to improve their adaptability across various datasets and tasks \cite{zhang2024graphtranslator, tang2024graphgpt}:
 \begin{equation}
    Y = f_\text{PLM}(\textsc{Prompt}, G_\text{seq}, f_\text{GNN}(\mathcal{G})).
\end{equation}

\subsection{Discussion}

While the message passing framework in classic GNNs excels in learning from graph structures, they face challenges in heterophilic graphs due to inconsistent neighborhood label distributions. Heterophily-specific GNNs address this by extending non-local neighbors and refining message passing via adaptive aggregation or inter-layer fusion. However, they still rely heavily on graph topology and often underutilize rich semantic information from textual content, which is crucial in distinguishing homophilic and heterophilic relationships.

In contrast, PLM-based methods can more effectively leverage the semantic features of textual content. However, they lack intrinsic awareness of graph structures and often resort to graph flattening techniques, which make them less effective in distinguishing homophilic and heterophilic structures. 
In addition, they often require computationally expensive fine-tuning for adaptation. 

Finally, co-training methods attempt to bridge the gap in GNN- and PLM-based methods by integrating GNNs with PLMs, thus benefiting from both graph structure and semantic features. However, their effective integration often depends on expensive fine-tuning, and most existing methods are not specifically targeted at heterophilic graphs. A recent study, LLM4HeG \cite{wu2024exploring}, achieves significant improvement on heterophilic graphs by incorporating specific designs to distinguish between heterophilic and homophilic neighbors based on semantic features. While it demonstrates the potential of PLMs for heterophilic graph learning, further research is needed to optimize information fusion strategies, enhance structural-semantic consistency, and improve computational efficiency in modeling heterophilic graphs.


%% file: exp.tex
\begin{table*}[!htb]
    \centering
    \small
    \caption{Accuracy benchmarking. The best results for each paradigm are highlighted in distinct colors.}
        \vspace{-2mm}
        \resizebox{\textwidth}{!}{%
    \begin{tabular}{c|c|c|ccccc}
        \toprule
        \multicolumn{2}{c|}{Paradigms}
        & Methods & Cornell & Texas & Wisconsin & Actor & Amazon \\
        \toprule
        \multirow{10}{*}{\makecell[c]{GNNs:\\ \textit{Shallow feature}}} & \multirow{3}{*}{\textit{w/o Heterophilic}}
        & GCN & 51.43$\pm$4.9 & 54.09$\pm$1.7 & 53.68$\pm$3.4 & 63.33$\pm$1.3 & 43.97$\pm$0.3 \\
        & & GraphSAGE & 71.43$\pm$1.9 & 68.18$\pm$0.2 & 71.35$\pm$4.1 & 71.09$\pm$0.4 & 47.98$\pm$0.3  \\
        & & GAT & 48.10$\pm$3.1 & 57.27$\pm$6.8 & 55.79$\pm$6.9 & 46.23$\pm$2.6 & 43.32$\pm$0.1 \\
        \cmidrule(lr){2-8}
        & \multirow{7}{*}{\textit{Heterophilic}} 
        & H2GCN & 65.71$\pm$1.9 & 79.09$\pm$2.7 & 79.30$\pm$1.3 & \textcolor{blue}{73.86$\pm$0.4}& \textcolor{blue}{50.46$\pm$0.3}  \\
        & & FAGCN & \textcolor{blue}{73.33$\pm$2.8} & \textcolor{blue}{82.50$\pm$5.0} & 80.35$\pm$2.1 & 73.79$\pm$0.6 & 47.50$\pm$0.4 \\
        & & JacobiConv & 71.64$\pm$1.3 & 73.41$\pm$4.9 & 74.20$\pm$5.4 & 72.44$\pm$1.2 & 48.77$\pm$0.2  \\
        & & GBK-GNN & 67.62$\pm$2.4 & 79.55$\pm$2.5 & 71.93$\pm$2.2 & 72.10$\pm$0.6 & 47.21$\pm$0.6  \\
        & & OGNN & 66.67$\pm$3.0 & 75.91$\pm$2.3 & 77.54$\pm$2.3 & 73.54$\pm$1.3 & 50.24$\pm$1.6  \\
        & & SEGSL & \textcolor{blue}{73.33$\pm$3.9} & 80.00$\pm$1.9 & \textcolor{blue}{81.05$\pm$1.4} & 71.36$\pm$0.3 &  46.69$\pm$0.4 \\
        & & DisamGCL & 56.67$\pm$3.1 & 65.00$\pm$1.2 & 58.60$\pm$2.0 & 70.60$\pm$0.2 & 44.37$\pm$0.1  \\
        \midrule \midrule
        
        \multirow{10}{*}{\makecell[c]{GNNs:\\ \textit{LLM feature}}} & \multirow{3}{*}{\textit{w/o Heterophilic}} 
        & GCN & 52.86$\pm$1.8 & 43.64$\pm$3.3 & 41.40$\pm$1.8 & 66.70$\pm$1.3 & 39.33$\pm$1.0 \\
        & & GraphSAGE & 75.71$\pm$1.8 & 81.82$\pm$2.5 & 80.35$\pm$1.3 & 70.37$\pm$0.1 & 46.63$\pm$0.1 \\
        & & GAT & 54.28$\pm$5.1 & 51.36$\pm$2.3 & 50.53$\pm$1.7 & 63.74$\pm$6.7 & 35.12$\pm$6.4 \\
        \cmidrule(lr){2-8}
        & \multirow{7}{*}{\textit{Heterophilic}} 
        & H2GCN & 69.76$\pm$3.0 & 79.09$\pm$3.5 & 80.18$\pm$1.9 & 70.73$\pm$0.9 & 47.09$\pm$0.3 \\
        & & FAGCN & \textcolor{red}{76.43$\pm$3.1} & 84.55$\pm$4.8 & \textcolor{red}{83.16$\pm$1.4} & \textcolor{red}{75.58$\pm$0.5} & \textcolor{red}{49.83$\pm$0.6} \\
        & & JacobiConv & 73.57$\pm$4.3 & 81.80$\pm$4.1 & 76.31$\pm$1.3 & 73.81$\pm$0.3 & 49.43$\pm$0.5 \\
        & & GBK-GNN & 66.19$\pm$2.8 & 80.00$\pm$3.0 & 72.98$\pm$3.3 & 72.49$\pm$1.0 & 44.90$\pm$0.3 \\
        & & OGNN & 71.91$\pm$1.8 & \textcolor{red}{85.00$\pm$2.3} & 79.30$\pm$2.1 & 72.08$\pm$2.4 & 47.79$\pm$1.6 \\
        & & SEGSL & 66.67$\pm$4.1 & \textcolor{red}{85.00$\pm$2.0} & 79.30$\pm$1.8 & 72.73$\pm$0.8 & 47.38$\pm$0.2 \\
        & & DisamGCL & 50.48$\pm$2.0 & 65.00$\pm$1.2 & 57.89$\pm$0.0 & 67.78$\pm$0.3 & 43.90$\pm$0.4 \\
        \midrule \midrule
        
        \multirow{7}{*}{PLMs} & \multirow{3}{*}{\textit{w/o Fine-tuning}} 
        & Bloom-560M & 5.24$\pm$2.8 & 6.82$\pm$1.4 & 7.02$\pm$3.1 & 0.32$\pm$0.2 & 0.03$\pm$0.1 \\
        & & Bloom-1B & 10.52$\pm$1.5 & 5.29$\pm$2.1 & 8.28$\pm$2.5 & 0.34$\pm$0.5 & 0.24$\pm$0.3 \\
        & & Vicuna-7B & 23.26$\pm$3.2 & 27.03$\pm$1.9 & 25.35$\pm$3.7 & 23.26$\pm$1.2 & 35.89$\pm$0.6 \\
        \cmidrule(lr){2-8}
        & \multirow{4}{*}{\textit{Fine-tuned}} 
        & Bloom-560M & 21.90$\pm$3.5 & 34.55$\pm$9.8 & 44.91$\pm$3.3 & 65.13$\pm$1.1 & 57.85$\pm$0.4 \\        
        & & Bloom-1B & 27.14$\pm$4.9 & 46.36$\pm$1.8 & 34.39$\pm$5.0 & 64.21$\pm$1.0 & 55.44$\pm$0.3 \\
        & & Vicuna-7B & \textcolor{cyan}{63.45$\pm$2.1} & 53.38$\pm$3.6 & \textcolor{cyan}{78.91$\pm$2.8} & \textcolor{cyan}{76.34$\pm$0.8} & \textcolor{cyan}{63.00$\pm$0.3} \\
        & & LLaGA & 51.27$\pm$2.4 & \textcolor{cyan}{58.32$\pm$1.9} & 43.24$\pm$3.0 & 72.09$\pm$1.3 & 38.24$\pm$0.7 \\
        \midrule \midrule
        
        \multirow{3}{*}{Co-training} & \multirow{2}{*}{\textit{w/o Heterophilic}} 
        & G2P2 & 43.81$\pm$1.2 & 53.64$\pm$1.8 & 56.14$\pm$4.6 & 61.73$\pm$0.3 & 43.64$\pm$0.3 \\
        & & GraphGPT & 30.32$\pm$2.0 & 42.36$\pm$1.4 & 23.58$\pm$2.0 & 40.29$\pm$0.9 & 41.20$\pm$0.1 \\
        
        \cmidrule(lr){2-8}
        & \multirow{1}{*}{\textit{Heterophilic}} 
        & LLM4HeG & \textcolor{magenta}{77.62$\pm$2.9} & \textcolor{magenta}{89.09$\pm$3.3} & \textcolor{magenta}{86.14$\pm$2.1} & \textcolor{magenta}{76.82$\pm$0.5} & \textcolor{magenta}{51.53$\pm$0.4} \\
        \bottomrule
    \end{tabular}
    \label{tab: result}}
    \vspace{-0.3cm}
\end{table*}

\section{Benchmarking Results and Analysis}
In this section, we evaluate the performance of existing methods on HeTGB and conduct a systematic analysis. More experiments on existing large-scale datasets with relatively low homophily ratios (e.g., Arxiv and Children \cite{yan2023comprehensive}) are provided in Appendix \ref{app:largescale}. 

\subsection{Experimental Setup}
\label{exp:setup}
\vspace{+0.1cm}\noindent\textbf{Baselines.} We conduct experiments on three learning paradigms. 
(1) \textbf{GNN-based methods}: the classic GCN \cite{kipf2016semi}, GraphSAGE \cite{hamilton2017inductive}, and GAT \cite{velivckovic2018graph}, as well as heterophily-specific H2GCN \cite{zhu2020beyond}, FAGCN \cite{bo2021beyond}, JacobiConv  \cite{wang2022powerful}, GBK-GNN \cite{du2022gbk}, OGNN \cite{songordered}, SEGSL \cite{zou2023se}, and DisamGCL \cite{zhao2024disambiguated}. (2) \textbf{PLM-based methods}: Bloom-560M \cite{le2023bloom}, Bloom-1B \cite{le2023bloom}, Vicuna-7B \cite{zheng2024judging} and LLaGA \cite{chenllaga}. (3) \textbf{Co-training methods}: G2P2 \cite{wen2023augmenting} and GraphGPT \cite{tang2024graphgpt} for general TAGs, as well as LLM4HeG \cite{wu2024exploring} for heterophilic TAGs. Additional experiments involving Transformer-based models are provided in Appendix \ref{app:transformer}. More details of baselines and implementation can be found in Appendix \ref{app:baselines} and \ref{app:prompt}, respectively. We report Accuracy as the primary metric in Table~\ref{tab: result}. To provide a more comprehensive evaluation under class imbalance, we additionally report Macro-F1 and Balanced Accuracy for representative methods in Appendix~\ref{app:additional_metrics}. All reported experiments are repeated ten times, with mean and standard deviation reported.




\vspace{-0.2cm}
\subsection{Analysis of GNN-based Methods}

To evaluate the ability of GNN-based methods to leverage semantic features in addition to graph structures, we compare the performance of shallow features and LLM-derived features across different GNN-based methods, as shown in Table~\ref{tab: result}. 
The results indicate that LLM-derived features generally lead to improved node classification accuracy, particularly for heterophily-specific GNNs. This highlights the advantage of leveraging richer semantic representations from textual node content in differentiating homophilic and heterophilic neighbors. 
However, the effectiveness of LLM-derived features is not consistent across all GNNs or all datasets, where we may observe a slight performance decrease compared to shallow features in some cases (\textit{e.g.}, H2GCN on Amazon). This variability arises because GNNs aggregate node embeddings into coarse-grained, fixed-dimensional representations, limiting their ability to capture fine-grained semantic knowledge even when enhanced by LLM-derived features. Consequently, while incorporating LLM-derived features can provide additional semantic contexts, it does not always lead to improved performance particularly when structural relationships play a more critical role.  

Moreover, compared to classic GNNs, heterophily-specific GNNs achieve overwhelmingly better performance, regardless of whether shallow or LLM-derived features are used. This can be attributed to heterophily-specific designs, including redefining the neighborhood in message passing and adjusting message aggregation to better accommodate heterophilic structures.


\vspace{-0.2cm}
\subsection{Analysis of PLM-based Methods}
On one hand, PLM-based methods without fine-tuning perform poorly, exhibiting limited generalization to graph data and tasks. This is not surprising, as PLMs are pre-trained on plain text corpora, making them unaware of structural dependencies for effective node classification in heterophilic graphs. On the other hand, fine-tuning PLMs often results in a significant performance boost. 
It indicates that when sufficiently fine-tuned, LLMs can effectively utilize task-relevant semantic patterns in node text, making them a promising solution for heterophilic graphs. Nevertheless, heterophily-specific GNNs with LLM-derived features tend to outperform fine-tuned LLMs, particularly when the homophily ratios are lower, highlighting the importance of structural information in heterophilic graph learning. 
These findings emphasize the need for task-specific adaptation for LLMs and motivate future research into hybrid co-training methods that integrate message passing in GNNs with LLM-driven reasoning to further enhance performance.


\subsection{Analysis of Co-training Methods}

To evaluate the effectiveness of co-training methods, we compare the performance of heterophily-specific and non-heterophilic approaches, as shown in Table~\ref{tab: result}. The results reveal a clear performance gap between general co-training methods and those specifically designed for heterophilic graphs. 
The non-heterophilic methods struggle to generalize in heterophilic settings, likely due to their lack of explicit adaptation to heterophilic relationships. In particular, they often underperform fine-tuned PLMs, suggesting that directly incorporating structural information without differentiating their homophilic and heterophilic characteristics may be harmful. 
In contrast, LLM4HeG \cite{wu2024exploring} significantly outperforms other co-training methods. This highlights the effectiveness of leveraging both PLMs and GNNs in a heterophily-aware manner, allowing for more effective message passing and representation learning in heterophilic graphs. However, LLM4HeG does not explicitly consider the alignment between graph structures and semantic features \cite{wen2023augmenting}, which could further enhance their integration. 

Our findings underscore the importance of integrating both structural and textual signals, suggesting that future research could benefit from heterophilic adaptations that enable better alignment between structure- and PLM-based representations. 

\vspace{-0.2cm}
\subsection{Cross-dataset Analysis}

To further understand how different learning paradigms behave across datasets, we compare their relative performance under different graph and text characteristics, as shown in Table~\ref{tab: result}. The results show that no single learning paradigm consistently performs best across all datasets. On the strongly heterophilic WebKB datasets, structure-aware GNNs generally show clear advantages over text-only PLMs, while on Actor and Amazon, fine-tuned PLMs become more competitive and can outperform several GNN-based and co-training methods. This suggests that the relative importance of structural and textual information varies across datasets, and different learning paradigms may be favored under different data characteristics.

Moreover, the performance does not vary monotonically with the graph homophily ratio. For example, both classic and heterophily-specific GNNs perform considerably worse on Amazon than on several WebKB datasets, although Amazon has a higher homophily ratio. This indicates that homophily is insufficient to characterize the difficulty of heterophilic TAG learning, and other dataset properties may also affect model performance.

We further observe that text length does not show a clear relationship with classification performance. For instance, fine-tuned PLMs achieve comparable performance on Wisconsin and Actor despite their substantially different text lengths. These observations suggest that model performance is jointly influenced by structural and textual characteristics rather than by a single dataset property.

\subsection{Efficiency Analysis}
\begin{figure}[tbp] 
\centering
    \centering
    \includegraphics[scale=0.45]{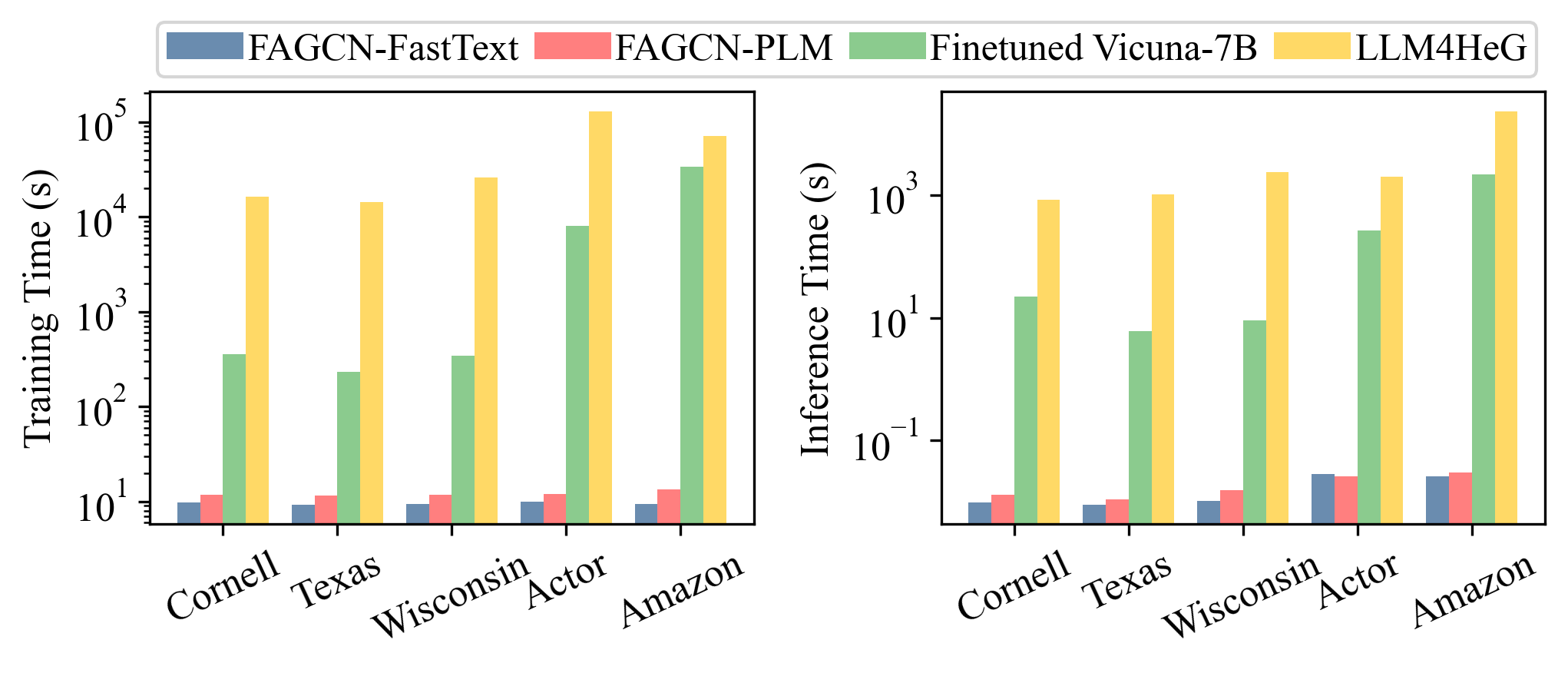}
\vspace{-0.6cm}
\caption{Training and inference times of the best-performing methods across different paradigms.}
\label{fig:traing and inference time}
\vspace{-0.5cm}
\end{figure}

To further evaluate the computational efficiency of different learning paradigms, we compare the training and inference time of the best-performing method within each paradigm, as shown in Figure~\ref{fig:traing and inference time}. GNN-based methods are the most efficient, although using LLM-derived features introduces a moderate overhead compared to shallow features due to more expensive feature processing. In contrast, both PLM-based and co-training methods incur substantially higher computational costs than GNN-based methods. Co-training methods incur the highest costs in both training and inference due to the need to train or fine-tune both PLMs and GNNs. While they achieve strong performance, their computational demands pose challenges for scalability in large-scale graph applications. These results emphasize the trade-off between model expressiveness and efficiency, motivating future research into more efficient co-training approaches without excessive computational overhead.

%% file: app.tex

\section{Distribution and Statistics}
\label{app:distribution}

Unlike previous heterophilic graph datasets that primarily rely on shallow embeddings derived from textual or numerical features, HeTGB utilizes raw text collected directly from original data sources. This results in expected variations in key statistics such as the number of nodes and edges, as shown in Table~\ref{tab: dataset}. As HeTGB is specifically designed for the investigation of heterophilic structures enriched with rich textual information, we further examine the distribution of the textual content and heterophilic contexts associated with nodes. 

\begin{figure*}[htbp] 
\centering
\subfigure[Cornell]{
    \centering
    \includegraphics[scale=0.4]{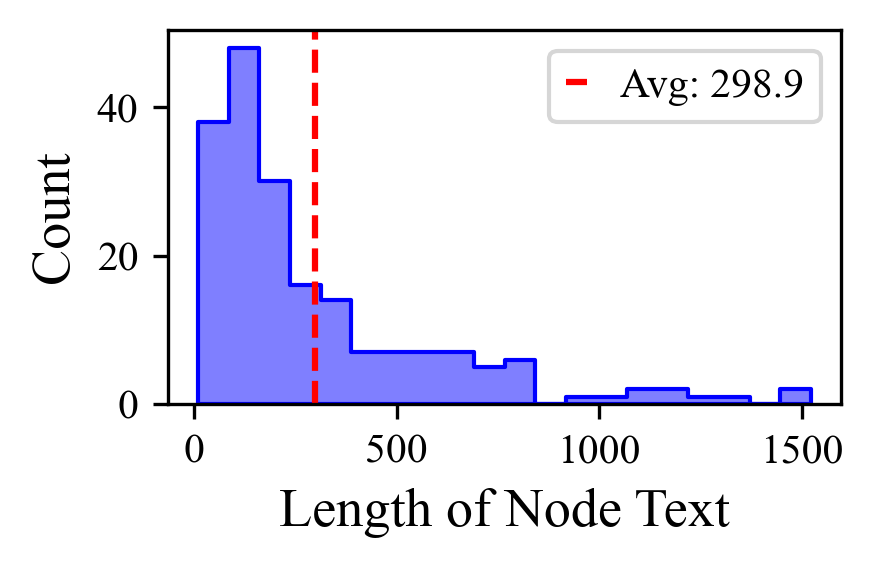}}
\subfigure[Texas]{
    \centering
    \includegraphics[scale=0.4]{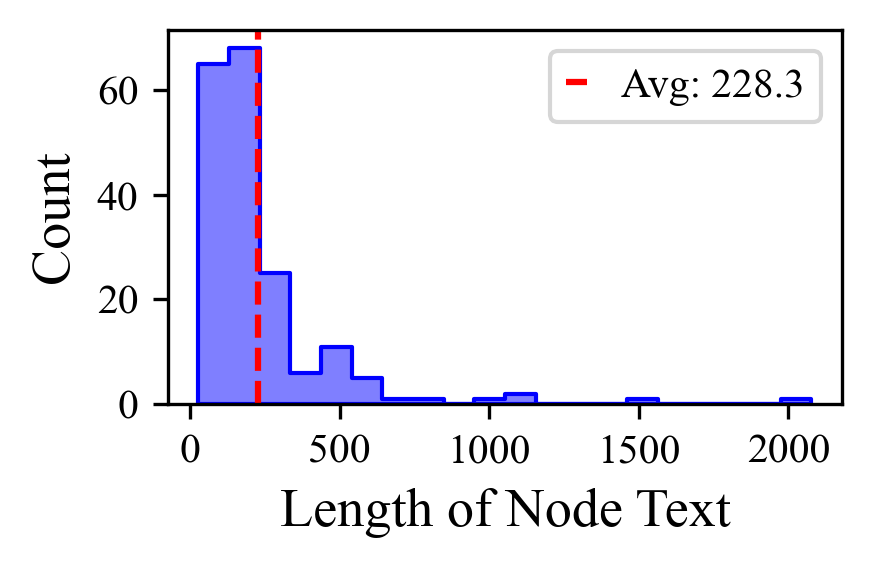}}
\subfigure[Wisconsin]{
    \centering
    \includegraphics[scale=0.4]{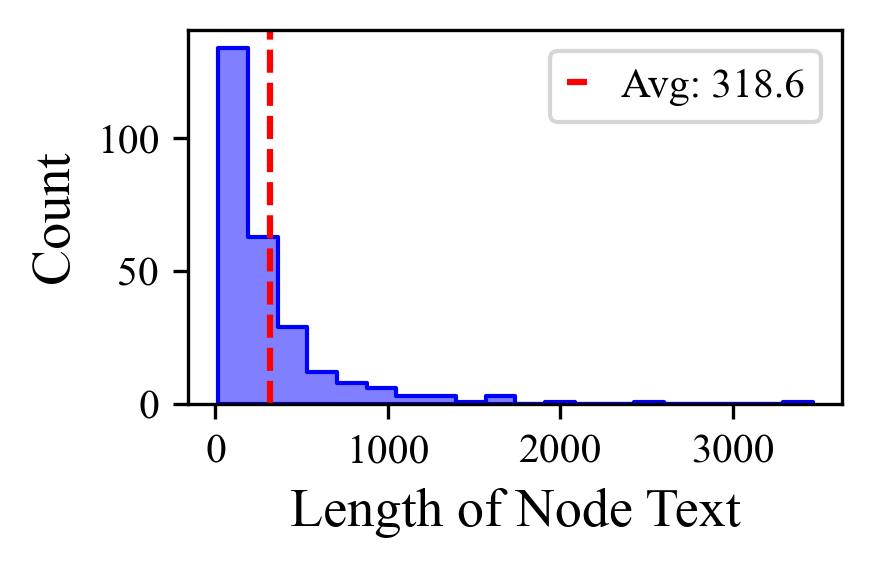}}
\subfigure[Actor]{
    \centering
    \includegraphics[scale=0.4]{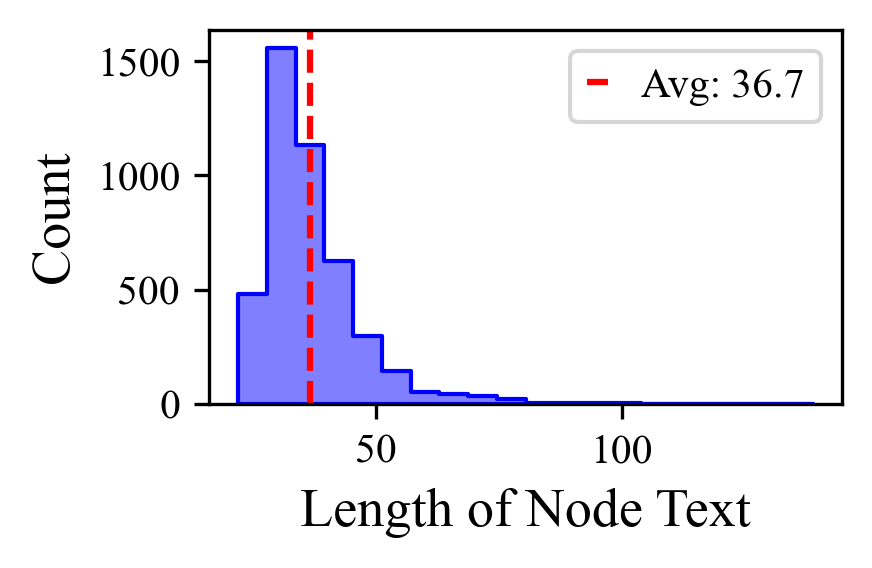}}
\subfigure[Amazon]{
    \centering
    \includegraphics[scale=0.4]{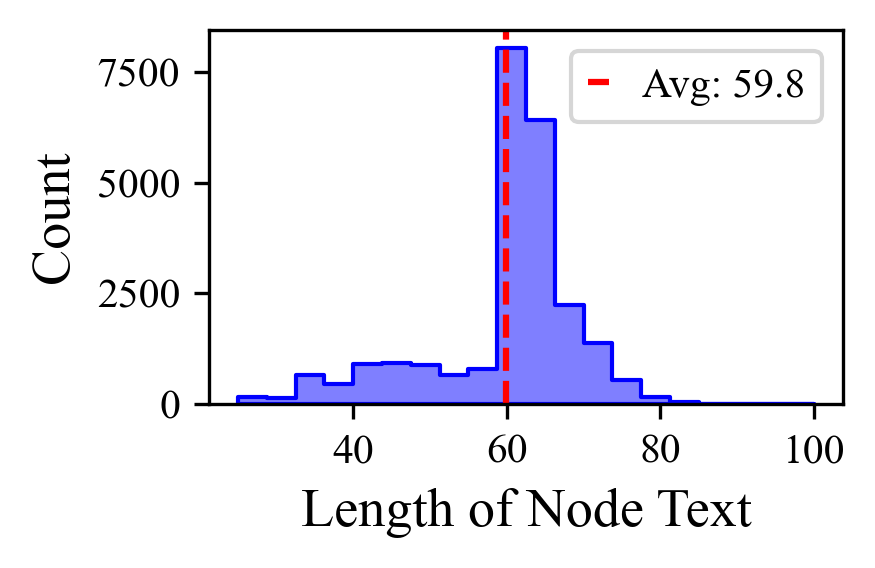}}
 \vspace{-2mm}
 \caption{Distribution of node text lengths, in terms of the number of tokens in each node's textual description.}
 \label{fig: textlen}
 \vspace{-2mm}
\end{figure*}

First, we analyze the distribution of node text lengths in Figure~\ref{fig: textlen}. The distributions highlight the varying extent of textual information associated with nodes, ranging from shorter descriptions in the Actor dataset to more extensive textual content in the Wisconsin and Cornell datasets. Notably, the Cornell, Texas, and Wisconsin datasets exhibit long-tail distributions, characterized by a few nodes with significantly longer text compared to the majority. This is because the raw text consists of entire web pages, which vary in length depending on the individual authors.
In contrast, the Actor and Amazon datasets display distributions closer to Gaussian, with text lengths concentrated around the mean, as indicated by the red dashed lines representing the average. This is because their raw text is manually extracted from original pages or user reviews, leading to a more consistent text length distribution. The variation in text lengths and the use of different text extraction or synthesis processes allow for a diverse benchmark for evaluating models' ability to handle heterophilic graphs with varying degrees of textual complexity.

\begin{figure*}[tbp] 
\centering
\subfigure[Cornell]{
    \centering
    \includegraphics[scale=0.4]{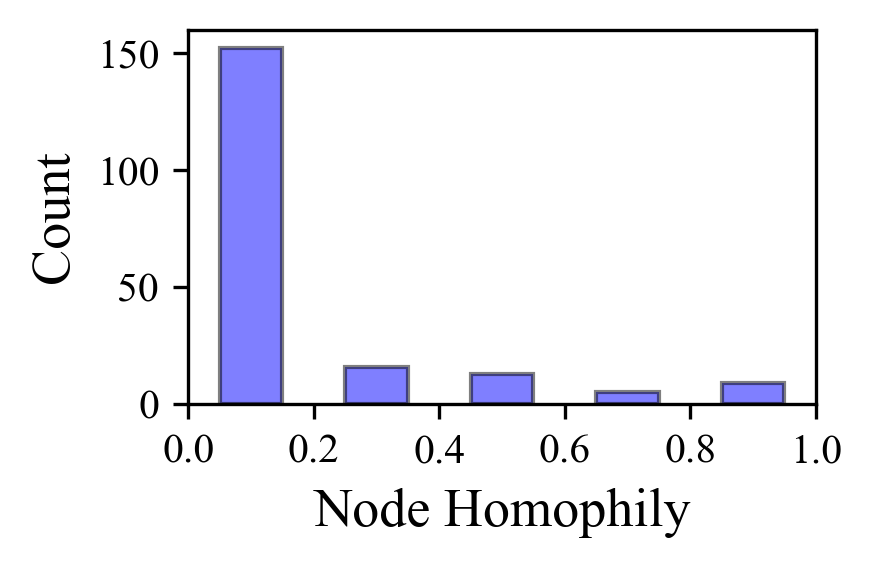}}
\subfigure[Texas]{
    \centering
    \includegraphics[scale=0.4]{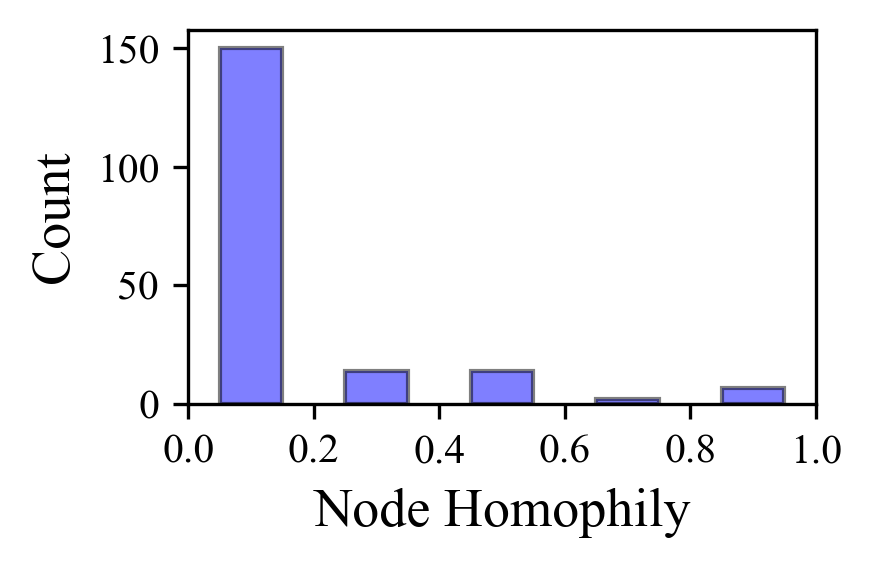}}
\subfigure[Wisconsin]{
    \centering
    \includegraphics[scale=0.4]{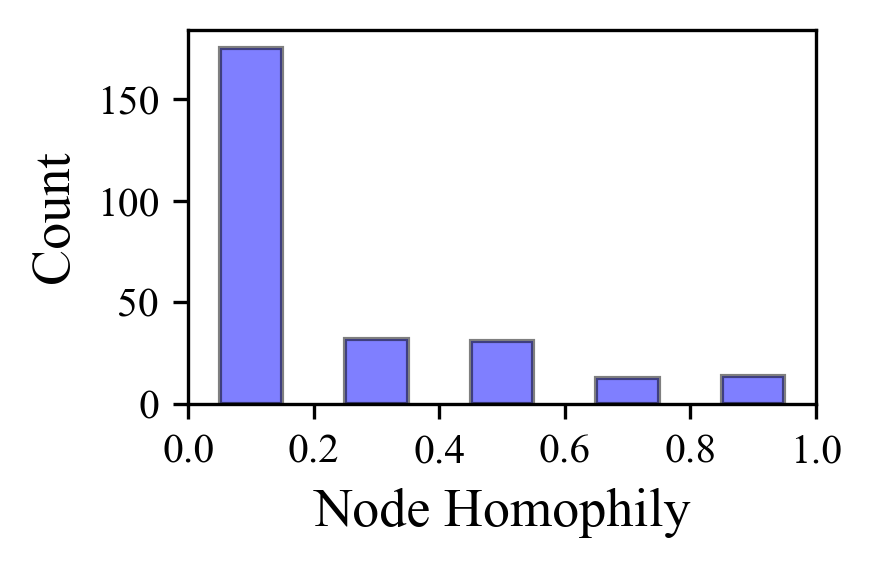}}
\subfigure[Actor]{
    \centering
    \includegraphics[scale=0.4]{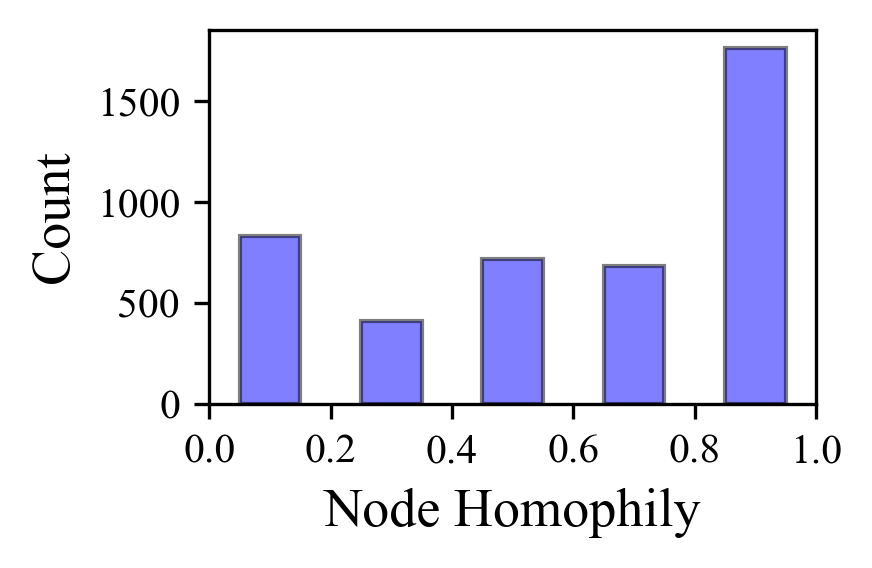}}
\subfigure[Amazon]{
    \centering
    \includegraphics[scale=0.4]{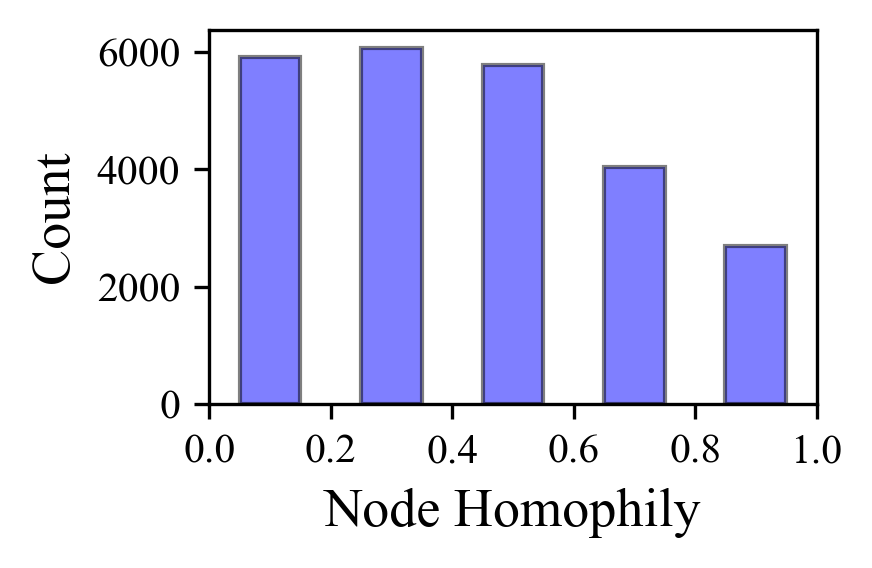}}
    \vspace{-2mm}
 \caption{Distribution of node homophily ratios.}
 \label{fig: nodehomo}
 \vspace{-3mm}
\end{figure*}

Additionally, we present the distribution of node homophily ratios in Figure~\ref{fig: nodehomo}, offering deeper insights into the heterophily characteristics of our datasets. 
In HeTGB, each dataset is characterized by heterophilic graph structures and rich text. A heterophilic graph is characterized by its low \emph{homophily ratio} indicating that linked nodes are more likely to have dissimilar features and belong to different classes. The homophily ratio can be defined based on edges \cite{zhu2020beyond}, representing the proportion of edges connecting nodes with the same class label across the entire graph. Given a graph $\mathcal{G}=(\mathcal{V},\mathcal{E})$ with a set of nodes $\mathcal{V}$ and edges $\mathcal{E}$, the edge homophily ratio $\mathcal{H}_{\text{e}}$ is computed as
\begin{align}
\mathcal{H}_{\text{e}}(\mathcal{G}) &= \textstyle \frac{|\{(v, u) \in \mathcal{E} : y_v = y_u\}|}{|\mathcal{E}|},
\label{eq:edge_homo}
\end{align}
where $y_v$ denotes the label of node $v$.

While the edge homophily ratio in Eq.~\eqref{eq:edge_homo}, $\mathcal{H}_e$, describes the global homophily characteristics of the entire graph, the node homophily ratio \cite{peigeom}, $\mathcal{H}_{\text{n}}(v,\mathcal{G})$, captures the local homophily pattern at each node $v$ in a graph $\mathcal{G}=(\mathcal{V},\mathcal{E})$:
\begin{align}
\mathcal{H}_{\text{n}}(v,\mathcal{G})&= \textstyle \frac{|\{u \in \mathcal{N}(v) : y_v = y_u\}|}{|\mathcal{N}(v)|}. 
\end{align}
where $\mathcal{N}(v)$ denotes the set of neighbors of node $v$. The Cornell, Texas, and Wisconsin datasets exhibit highly skewed distributions across nodes with a strong presence of low-homophily nodes, reinforcing their heterophilic nature. 
In contrast, the Actor and Amazon datasets display a more mixed distribution, capturing both homophilic and heterophilic patterns across nodes. 
These observations validate that our datasets are characterized by diverse heterophilic graph structures, enabling robust evaluation of models designed for heterophilic TAGs.

\section{Data Integrity, Privacy, and Label Leakage Analysis}
\label{app:data_integrity}

\paragraph{Data sources and privacy.}
HeTGB is reconstructed entirely from publicly accessible data sources without collecting new data from human participants. Cornell, Texas, and Wisconsin are reconstructed from the original WebKB webpages\footnote{\url{https://www.cs.cmu.edu/afs/cs.cmu.edu/project/theo-11/www/wwkb/}}, with their graph structures and URLs obtained from a public repository\footnote{\url{https://github.com/sqrhussain/structure-in-gnn/tree/master/data/graphs/raw/WebKB}}. Actor is reconstructed from the publicly released film-director-actor-writer network\footnote{\url{https://www.aminer.cn/lab-datasets/soinf/}}, whose textual information is derived from Wikipedia. Amazon is constructed from the Amazon product co-purchasing metadata provided by the Stanford Large Network Dataset Collection\footnote{\url{https://snap.stanford.edu/data/amazon-meta.html}}. Since the textual attributes originate from real-world public sources, they may retain publicly available person-related or user-contributed information. Our preprocessing focuses on benchmark reconstruction and the removal of task-specific label information rather than exhaustive de-identification or content moderation.

\paragraph{Preprocessing against direct label leakage.}
We explicitly remove information that directly reveals the prediction targets when constructing the textual attributes. For Cornell, Texas, and Wisconsin, node text is extracted from the original WebKB webpages rather than generated from class labels. For Actor, category terms corresponding to the five prediction labels are removed from the raw text. For Amazon, we remove the final average product rating, which defines the prediction label, and retain only product descriptions and up to five early individual ratings.

\paragraph{Residual label leakage analysis.}
For Cornell, Texas, and Wisconsin, we examine occurrences of the class names \textit{student}, \textit{course}, \textit{project}, \textit{staff}, and \textit{faculty} in the raw webpage text. In the three datasets, respectively, 24.6\%, 40.1\%, and 35.5\% of nodes contain none of these terms; 47.2\%, 30.5\%, and 32.1\% contain at least one class-name term inconsistent with the ground-truth class; and 30.3\%, 17.1\%, and 18.9\% contain multiple class-name terms. Thus, these naturally occurring terms do not provide a deterministic lexical mapping to the labels. These statistics are not mutually exclusive, as a webpage may satisfy more than one of these conditions.
For Actor, after removing explicit category terms, we further audit indirect lexical cues associated with nationality and acting domains. Among all 4,416 nodes, 83.4\% contain no detected cues plausibly associated with the ground-truth category, 13.9\% contain potentially consistent cues, and 2.7\% contain cues associated with a different category. Hence, most Actor instances cannot be classified through straightforward matching of such indirect cues.
For Amazon, we measure the dependence between the retained early ratings $X$ and the final rating category $Y$ using normalized mutual information $I(X;Y)/H(Y)$. The resulting value is 21.3\%, indicating that early ratings provide partial label-related information but do not deterministically reveal the target class.
Taken together, these analyses suggest that, although the textual attributes may contain naturally occurring label-related cues, they do not trivially determine the ground-truth labels.

\section{Evaluation on Existing Benchmarks}
\label{app:largescale}

In this section, we further conduct experiments on existing large-scale datasets with relatively low homophily ratios, such as Arxiv and Children \cite{yan2023comprehensive}. This setting complements our benchmark by introducing larger graphs with different structural and semantic distributions. We evaluate representative methods across three learning paradigms: GNNs (with both shallow and LLM-derived features), PLMs (zero-shot and fine-tuned), and co-training methods. The results are summarized in Table~\ref{tab:appendix-results}. We observe consistent patterns with our main benchmark: LLM-derived features tend to improve GNN performance; fine-tuned PLMs outperform zero-shot variants; and co-training methods such as G2P2 and LLM4HeG maintain strong performance across datasets. These findings further support the applicability of the studied methods beyond the proposed benchmark datasets.

\vspace{-0.2cm}
\begin{table}[h]
\centering
\small
\caption{Accuracy on Arxiv and Children datasets.}
\vspace{-0.2cm}
\label{tab:appendix-results}
\setlength{\tabcolsep}{5pt}
\begin{tabular}{llcc}
\toprule
\textbf{Paradigms} & \textbf{Method} & \textbf{Arxiv} & \textbf{Children} \\
\midrule
GNNs & GraphSAGE     & 73.51$\pm$0.2 & 57.84$\pm$0.5 \\
\scriptsize(Shallow features)   & FAGCN         & 64.72$\pm$0.5 & 44.91$\pm$0.5 \\
\midrule
GNNs & GraphSAGE     & 73.64$\pm$0.5 & 58.40$\pm$0.6 \\
\scriptsize(LLM features)  & FAGCN         & 71.89$\pm$0.8 & 53.70$\pm$0.7 \\
\midrule
PLMs {\scriptsize w/o Fine-tuning}   & Vicuna-7B       & 4.32$\pm$0.8           & 5.73$\pm$0.3           \\
\midrule
PLMs {\scriptsize Fine-tuned}      & Vicuna-7B       & 73.80$\pm$0.9          & 59.46$\pm$0.6          \\
\midrule
\multirow{2}{*}{Co-training} 
  & G2P2          & 79.84$\pm$1.3 & 55.91$\pm$1.2 \\
  & LLM4HeG       & 72.79$\pm$0.8 & 56.46$\pm$0.7 \\
\bottomrule
\end{tabular}
\vspace{-0.2cm}
\end{table}

\section{Evaluation on Graph Transformers}
\label{app:transformer}

Graph Transformers have emerged as an alternative to message-passing GNNs for graph representation learning. By leveraging self-attention mechanisms, these models can capture interactions between distant nodes and model long-range dependencies, which can be beneficial for heterophilic graphs, where informative signals may not lie in immediate neighborhoods.

To further broaden our evaluation, we additionally include a graph transformer baseline, CoBFormer \cite{xing2024less}, which addresses the over-globalizing problem in Graph Transformers by introducing a bi-level architecture with inter-cluster and intra-cluster attention. Consistent with the feature settings used for other models, we evaluate CoBFormer with both shallow and LLM-derived features. We conduct experiments on both the datasets in our proposed HeTGB benchmark and the additional datasets (Arxiv and Children) used in Appendix \ref{app:largescale}. 

The results show that incorporating LLM-derived textual features generally improves the performance of CoBFormer, yielding gains on six of the seven datasets, while a slight decrease is observed on Cornell. Overall, CoBFormer achieves performance comparable to heterophily-specific GNN baselines. These observations further highlight the potential value of rich textual semantics when learning on heterophilic graphs.

\vspace{-0.2cm}
\begin{table}[h]
\centering
\small
\caption{Accuracy of CoBFormer on HeTGB datasets and additional benchmarks.}
\vspace{-0.2cm}
\label{tab:cobformer-rotated}
\setlength{\tabcolsep}{12pt} 
\begin{tabular}{lcc}
\toprule
\textbf{Dataset} 
  & \makecell[c]{\textbf{CoBFormer} \\ \scriptsize(Shallow features)} 
  & \makecell[c]{\textbf{CoBFormer} \\ \scriptsize(LLM features)} \\
\midrule
Cornell      & 66.19$\pm$3.9  & 64.33$\pm$2.1 \\
Texas        & 73.82$\pm$4.4  & 81.82$\pm$3.6 \\
Wisconsin    & 77.54$\pm$1.9  & 78.95$\pm$2.5 \\
Actor        & 73.45$\pm$0.5  & 73.86$\pm$1.0 \\
Amazon       & 49.25$\pm$0.7  & 52.56$\pm$1.0 \\
Arxiv        & 69.70$\pm$0.4  & 71.30$\pm$0.3 \\
Children     & 52.28$\pm$0.5  & 54.07$\pm$0.4 \\
\bottomrule
\end{tabular}
\vspace{-0.2cm}
\end{table}

\section{Descriptions of Baselines}
\label{app:baselines}

\begin{itemize}[noitemsep,topsep=0pt]
\item GCN 
\cite{kipf2016semi}, GraphSAGE
\cite{hamilton2017inductive} and GAT
\cite{velivckovic2018graph} are widely used classic GNNs that aggregate information from local neighborhoods to learn node representations.
    
\item H2GCN 
 \cite{zhu2020beyond} enhances heterophilic graph learning by incorporating higher-order neighbors, ego-neighbor embedding separation, and intermediate-layer representations. 

\item FAGCN 
\cite{bo2021beyond} employs a self-gating mechanism to adaptively integrate low- and high-frequency signals, enabling robust learning across both homophilic and heterophilic graphs. 

\item  JacobiConv 
\cite{wang2022powerful} eliminates nonlinearity and utilizes Jacobi polynomial bases for spectral filtering, improving flexibility and expressiveness in graph signal learning. 

\item GBK-GNN 
\cite{du2022gbk} applies a learnable kernel selection mechanism to differentiate homophilic and heterophilic node pairs, optimizing neighborhood aggregation. 

\item  OGNN 
\cite{songordered} introduces an ordered gating mechanism for message passing, enhancing node interactions while mitigating over-smoothing in heterophilic graphs. 

\item  SEGSL 
\cite{zou2023se} refines graph topology using structural entropy and encoding trees, improving robustness against noisy edges and adversarial attacks. 

\item  DisamGCL 
\cite{zhao2024disambiguated} employs topology-aware contrastive learning to disambiguate node embeddings, addressing representation challenges in heterophilic and noisy graphs. 

\item G2P2 
\cite{wen2023augmenting} enhances graph-text alignment via contrastive pre-training and prompt-based strategies, benefiting low-resource text classification tasks. 

\item GraphGPT 
\cite{tang2024graphgpt} aligns graph structures with LLMs using self-supervised and task-specific instruction tuning, improving graph reasoning capabilities. 

\item LLaGA 
\cite{chenllaga} restructures graph nodes into token-compatible sequences, enabling LLMs to process graph data while preserving general-purpose adaptability. 

\item LLM4HeG 
\cite{wu2024exploring} integrates LLMs into GNNs for heterophilic TAGs, using edge discrimination and adaptive edge reweighting to enhance node classification. 
\end{itemize}

\section{More Experimental Setup}
\label{app:prompt}

\vspace{+0.1cm}\noindent\textbf{Data splits and hyperparameters.}
Following standard practice in prior studies \cite{zhu2020beyond}, we randomly split the nodes into training, validation, and test sets with a ratio of 48\%/32\%/20\% for Cornell, Texas, Wisconsin, and Actor. For Amazon, we adopt a 50\%/25\%/25\% split, consistent with prior work \cite{platonovcritical}. The hyperparameters of each method are selected following or tuned based on their respective original papers.

\vspace{+0.1cm}\noindent\textbf{Implementation Details.} GNN-based methods are primarily implemented based on publicly available code repositories. 
To assess their effectiveness in leveraging different types of node representations, we benchmark them with initial node features derived from shallow models (FastText \cite{grave2018learning}) and LLMs (Vicuna-7B \cite{zheng2024judging}).  
For PLM-based methods, including Bloom-560M, Bloom-1B \cite{le2023bloom}, and Vicuna-7B \cite{zheng2024judging}, we design prompts that directly incorporate textual node attributes, following the structured template used in LLM4HeG \cite{wu2024exploring}.
Our experiments follow two strategies: (1) zero-shot inference, where we directly input the prompts and data into the PLMs for answer generation, and (2) fine-tuning with LoRA \cite{hulora} to better adapt to the node classification task on heterophilic graphs.
For the co-training methods, we use the official implementations provided by the respective authors.


\vspace{+0.05cm}\noindent\textbf{Prompt Template.}
As mentioned in \ref{method-plm}, we adopt a prompt without structural information to avoid introducing conflicting semantics from heterophilous neighbors \cite{chen2024exploring}. We adopt the prompt format used in LLM4HeG \cite{wu2024exploring} for PLM-based methods, including Bloom-560M, Bloom-1B \cite{le2023bloom}, and Vicuna-7B \cite{zheng2024judging}. An example prompt for the Cornell, Texas, and Wisconsin datasets is provided below. 

\vspace{+0.1cm}\noindent\framebox{\parbox{0.97\linewidth}{\small
\emph{Background:} I have a dataset containing web page information collected from computer science department websites of various universities. These web pages have been manually categorized into five categories, including student, staff, faculty, course, and project.\\

\emph{Task:} I will provide you with the text information of a web page, and I would like you to classify it into one of the following categories: student, staff, course, faculty, or project.\\

The web page content: <text>\\

You may only output the category name, and do not discuss anything else!
}}
\\[2mm]

For LLaGA \cite{chenllaga}, GraphGPT \cite{tang2024graphgpt}, and LLM4HeG \cite{wu2024exploring}, we follow the prompt design as outlined in the official formats provided by the respective authors. Below are the specific prompt formats for each method, using the Cornell, Texas, and Wisconsin datasets as examples.

\vspace{+0.2cm}\noindent LLaGA:

\vspace{+0.1cm}\noindent\framebox{\parbox{0.97\linewidth}{\small
Given a node-centered graph: <node sequence>, each node represents a web page, we need to classify the center node into one of the following categories: student, staff, course, faculty, or project.\\

Please tell me which class the center node belongs to.
}}\\[1mm]

\vspace{+0.1cm}\noindent GraphGPT:

\vspace{+0.1cm}\noindent\framebox{\parbox{0.97\linewidth}{\small
Given a node-centered graph: <graph>, where the 0th node is the target web page, with the following information: <text>\\

\emph{Question}: Each node represents a web page, and we need to classify the center node into one of the following categories: student, staff, course, faculty, or project.\\

Please tell me which class the center node belongs to.
}}\\[1mm]

\vspace{+0.1cm}\noindent LLM4HeG:

\vspace{+0.1cm}\noindent\framebox{\parbox{0.97\linewidth}{\small
\emph{Background:} I have a dataset containing web page information collected from computer science department websites of various universities. These web pages have been manually categorized into five categories, including student, staff, faculty, course, and project.\\

\emph{Task:} I will provide you with the content of two web pages, and I want you to determine if they belong to the same category among student, staff, course, faculty, and project.\\

The first web page: <text>\\
The second web page: <text>\\

\emph{Answer template}: Yes or No. Please think step by step.
}}\\[2mm]

For other datasets, we simply modify the dataset description and category information in the prompt. The rest of the prompt structure remains unchanged.

\section{Additional Evaluation Metrics}
\label{app:additional_metrics}

To complement the Accuracy results reported in the main text, we further evaluate representative methods using Macro-F1 and Balanced Accuracy. These metrics provide additional perspectives on per-class performance, particularly for datasets with imbalanced class distributions. For open-ended PLM-based methods and GraphGPT, whose generated outputs are not strictly constrained to the predefined label space, Macro-F1 is less directly comparable with closed-set classifiers. We therefore report Balanced Accuracy only for these methods.

\begin{table*}[t]
\centering
\small
\caption{Macro-F1 (\%) of representative methods on HeTGB. Results are reported as mean $\pm$ standard deviation over 10 runs.}
\label{tab:macro_f1}
\setlength{\tabcolsep}{5pt}
\begin{tabular}{llccccc}
\toprule
\textbf{Paradigm} & \textbf{Method} & \textbf{Cornell} & \textbf{Texas} & \textbf{Wisconsin} & \textbf{Actor} & \textbf{Amazon} \\
\midrule

\multirow{3}{*}{\shortstack[l]{GNNs: \\ Shallow feature}}
& GCN   & 39.31$\pm$6.3 & 50.25$\pm$2.4 & 42.37$\pm$3.6 & 38.29$\pm$1.2 & 27.14$\pm$0.4 \\
& H2GCN & 44.27$\pm$2.4 & 62.90$\pm$4.3 & 58.40$\pm$1.6 & 55.93$\pm$0.8 & 30.49$\pm$0.5 \\
& FAGCN & 57.09$\pm$1.6 & 63.57$\pm$8.3 & 64.63$\pm$1.9 & 53.08$\pm$0.7 & 26.96$\pm$1.1 \\
\midrule

\multirow{3}{*}{\shortstack[l]{GNNs: \\ LLM feature}}
& GCN   & 41.29$\pm$4.3 & 52.83$\pm$4.9 & 41.26$\pm$3.0 & 42.29$\pm$1.1 & 29.20$\pm$0.4 \\
& H2GCN & 58.29$\pm$3.9 & 61.29$\pm$3.9 & 65.27$\pm$1.0 & 50.29$\pm$0.8 & 27.05$\pm$1.1 \\
& FAGCN & 59.93$\pm$6.8 & 68.89$\pm$3.4 & 63.29$\pm$5.3 & 57.63$\pm$0.3 & 31.85$\pm$1.1 \\
\midrule

Co-training
& LLM4HeG & 63.45$\pm$1.1 & 71.12$\pm$3.7 & 67.17$\pm$2.3 & 61.58$\pm$0.9 & 32.71$\pm$0.8 \\

\bottomrule
\end{tabular}
\end{table*}

\begin{table*}[t]
\centering
\small
\caption{Balanced Accuracy (\%) of representative methods on HeTGB. Results are reported as mean $\pm$ standard deviation over 10 runs.}
\label{tab:balanced_acc}
\setlength{\tabcolsep}{5pt}
\begin{tabular}{llccccc}
\toprule
\textbf{Paradigm} & \textbf{Method} & \textbf{Cornell} & \textbf{Texas} & \textbf{Wisconsin} & \textbf{Actor} & \textbf{Amazon} \\
\midrule

\multirow{3}{*}{\shortstack[l]{GNNs: \\ Shallow feature}}
& GCN   & 40.25$\pm$5.9 & 49.13$\pm$2.3 & 44.26$\pm$2.7 & 35.24$\pm$2.9 & 29.15$\pm$0.9 \\
& H2GCN & 43.20$\pm$4.2 & 66.10$\pm$5.9 & 58.29$\pm$1.1 & 50.20$\pm$0.7 & 26.19$\pm$0.3 \\
& FAGCN & 57.65$\pm$1.2 & 63.94$\pm$7.7 & 64.97$\pm$2.7 & 50.56$\pm$0.6 & 30.00$\pm$0.6 \\
\midrule

\multirow{3}{*}{\shortstack[l]{GNNs: \\ LLM feature}}
& GCN   & 40.29$\pm$4.2 & 52.47$\pm$2.1 & 40.37$\pm$2.0 & 45.23$\pm$1.4 & 33.30$\pm$0.2 \\
& H2GCN & 59.36$\pm$5.2 & 64.58$\pm$2.7 & 67.13$\pm$1.4 & 49.20$\pm$0.3 & 30.22$\pm$0.3 \\
& FAGCN & 60.24$\pm$5.1 & 67.35$\pm$4.5 & 62.80$\pm$5.4 & 55.92$\pm$0.4 & 33.13$\pm$0.8 \\
\midrule

\multirow{2}{*}{\shortstack[l]{PLMs: \\ w/o Fine-tuning}}
& Bloom-560M & 4.79$\pm$3.1 & 3.98$\pm$1.5 & 7.65$\pm$3.9 & 0.43$\pm$0.4 & 0.09$\pm$0.9 \\
& Bloom-1B   & 10.56$\pm$2.5 & 2.91$\pm$2.4 & 10.50$\pm$4.3 & 0.33$\pm$0.4 & 0.18$\pm$0.6 \\
\midrule

\multirow{2}{*}{\shortstack[l]{PLMs: \\ Fine-tuned}}
& Bloom-560M & 11.11$\pm$1.9 & 15.07$\pm$4.6 & 24.33$\pm$1.7 & 52.38$\pm$0.7 & 51.57$\pm$0.5 \\
& Bloom-1B   & 12.89$\pm$2.2 & 19.18$\pm$1.4 & 21.04$\pm$4.8 & 52.13$\pm$1.1 & 48.60$\pm$0.4 \\
\midrule

\multirow{2}{*}{Co-training}
& GraphGPT & 15.09$\pm$1.4 & 18.27$\pm$1.0 & 16.09$\pm$3.3 & 33.86$\pm$2.1 & 33.47$\pm$0.4 \\
& LLM4HeG  & 61.30$\pm$1.3 & 68.50$\pm$4.5 & 64.50$\pm$1.4 & 54.17$\pm$0.4 & 34.59$\pm$0.2 \\

\bottomrule
\end{tabular}
\end{table*}

The additional metrics reveal performance differences that are less visible from Accuracy alone. For example, GCN with shallow features achieves 43.97\% Accuracy but only 27.14\% Macro-F1 on Amazon, indicating weaker class-wise performance. Nevertheless, the main benchmarking trends remain broadly consistent across metrics: heterophily-specific GNNs generally outperform vanilla GCN, while LLM-derived features provide gains in many settings. The noticeable variance of some PLM-based methods also highlights the importance of repeated evaluation.

\section{Analysis of Prompt Design}
\label{app:prompt_analysis}

To further investigate the effect of prompt design on PLM-based methods, we conduct a controlled study using zero-shot Vicuna-7B with four prompt configurations. Our default \textit{Text-only} prompt in Appendix~\ref{app:prompt} contains only the textual attribute of the target node and excludes structural information. We compare it with variants that introduce either explicit reasoning instructions or neighborhood information. Except for the prompt construction, all other experimental settings are kept unchanged.

\begin{table*}[!t]
\centering
\small
\caption{Effect of different prompt designs for Vicuna-7B on HeTGB. Results are reported as mean Accuracy (\%) $\pm$ standard deviation over 10 runs.}
\label{tab:prompt_design}
\setlength{\tabcolsep}{6pt}
\begin{tabular}{lccccc}
\toprule
\textbf{Prompt Design} & \textbf{Cornell} & \textbf{Texas} & \textbf{Wisconsin} & \textbf{Actor} & \textbf{Amazon} \\
\midrule
Text-only
& 23.26$\pm$3.2
& 27.03$\pm$1.9
& 25.35$\pm$3.7
& 23.26$\pm$1.2
& 35.89$\pm$0.6 \\

Text-only + CoT
& 23.10$\pm$3.0
& 27.86$\pm$2.2
& 26.18$\pm$3.4
& 22.45$\pm$1.1
& 37.31$\pm$0.7 \\

Raw neighbors
& 17.74$\pm$3.5
& 24.36$\pm$2.5
& 21.71$\pm$3.9
& 23.48$\pm$1.3
& 35.21$\pm$0.8 \\

Selected neighbors
& 25.31$\pm$3.1
& 28.14$\pm$2.0
& 27.42$\pm$3.3
& 24.67$\pm$1.1
& 35.02$\pm$0.7 \\
\bottomrule
\end{tabular}
\end{table*}

Specifically, we consider the following variants:
\begin{itemize}[noitemsep,topsep=0pt,leftmargin=*]
    \item \textbf{Text-only}: the default prompt used in our main experiments.
    
    \item \textbf{Text-only + CoT}: the same textual input as the default prompt, with an additional instruction to reason step by step before producing the final prediction.
    
    \item \textbf{Raw neighbors}: we randomly sample up to three one-hop neighbors of the target node and append their textual attributes to the prompt.
    
    \item \textbf{Selected neighbors}: we rank the one-hop neighbors according to the cosine similarity between their Vicuna-7B-derived textual representations and that of the target node, and retain the top three most relevant neighbors.
\end{itemize}

For both neighborhood-based variants, we use all available neighbors when the node degree is smaller than three, ensuring that the two settings differ only in how neighboring nodes are selected.

The results in Table~\ref{tab:prompt_design} show that PLM performance is sensitive to how textual and structural information is incorporated into the prompt. Adding chain-of-thought instructions leads to limited and inconsistent changes across datasets, suggesting that reasoning instructions alone do not consistently improve classification performance. In contrast, incorporating randomly sampled neighbors generally degrades performance on the strongly heterophilic WebKB datasets, indicating that unfiltered neighborhood information may introduce conflicting semantics. Selecting semantically relevant neighbors effectively mitigates this issue and improves over the text-only baseline on Cornell, Texas, Wisconsin, and Actor, although the improvement does not extend to Amazon. These results suggest that neighborhood information can benefit PLM-based methods when incorporated selectively, while directly introducing heterophilic neighbors may be detrimental.

\vspace{+0.1cm}\noindent\textbf{Prompt modifications.}
The prompt variants follow the same dataset descriptions and candidate label spaces as the default prompt in Appendix~\ref{app:prompt}. For \textit{Text-only + CoT}, we keep the original textual input unchanged and replace the final instruction with:

\begin{quote}
\small
Please analyze the webpage content step by step and determine the most likely category. In the final line, only output the category name.
\end{quote}

For the two neighborhood-based variants, we augment the default prompt with the target node and up to three neighboring nodes using the following format:

\begin{quote}
\small
The target web page: <target text>

The neighboring web pages: \\
<neighbor 1 text> \\
<neighbor 2 text> \\
<neighbor 3 text>

Note that neighboring web pages may belong to different categories from the target page.

You may only output the category name, and do not discuss anything else!
\end{quote}

The same neighbor-augmented prompt is used for both \textit{Raw neighbors} and \textit{Selected neighbors}; only the neighbor selection strategy differs. For other datasets, we modify only the dataset description and candidate categories accordingly.